\documentclass[sigconf]{aamas}  % do not change this line!

\usepackage{booktabs}
\usepackage{subfigure} 

\setcopyright{ifaamas}  % do not change this line!
\acmDOI{}  % do not change this line!
\acmISBN{}  % do not change this line!
\acmConference[AAMAS'18]{Proc.\@ of the 17th International Conference on Autonomous Agents and Multiagent Systems (AAMAS 2018)}{July 10--15, 2018}{Stockholm, Sweden}{M.~Dastani, G.~Sukthankar, E.~Andr\'{e}, S.~Koenig (eds.)}  % do not change this line!
\acmYear{2018}  % do not change this line!
\copyrightyear{2018}  % do not change this line!
\acmPrice{}  % do not change this line!

\begin{document}

\title{Learning Temporal Strategic Relationships using Generative Adversarial Imitation Learning}  % put your title here!
%\titlenote{Produces the permission block, and copyright information}

% AAMAS: as appropriate, uncomment one subtitle line; see camera ready instructions
%\subtitle{Extended Abstract}
%\subtitle{Industrial Applications Track}
%\subtitle{Socially Interactive Agents Track}
%\subtitle{Blue Sky Ideas Track}
%\subtitle{Robotics Track}
%\subtitle{JAAMAS Track}
%\subtitle{Doctoral Mentoring Program}

%\subtitlenote{The full version of the author's guide is available as \texttt{acmart.pdf} document}

\author{Tharindu~Fernando \hspace{1cm} Simon Denman \hspace{1cm} Sridha Sridharan \hspace{1cm} Clinton Fookes}
\affiliation{\institution{Image and Video Research Laboratory, Queensland University of Technology (QUT), Australia}}
\email{ t.warnakulasuriya, s.denman, s.sridharan, c.fookes @qut.edu.au}

% The example's default list of authors is too long for headers
\renewcommand{\shortauthors}{T. Fernando et al.}

\begin{abstract}  % put your abstract here!
This paper presents a novel framework for automatic learning of complex strategies in human decision making. The task that we are interested in is to better facilitate long term planning for complex, multi-step events. We observe temporal relationships at the subtask level of expert demonstrations, and determine the different strategies employed in order to successfully complete a task. To capture the relationship between the subtasks and the overall goal, we utilise two external memory modules, one for capturing dependencies within a single expert demonstration, such as the sequential relationship among different sub tasks,  and a global memory module for modelling task level characteristics such as best practice employed by different humans based on their domain expertise. Furthermore, we demonstrate how the hidden state representation of the memory can be used as a reward signal to smooth the state transitions, eradicating subtle changes. We evaluate the effectiveness of the proposed model for an autonomous highway driving application, where we demonstrate its capability to learn different expert policies and outperform state-of-the-art methods. The scope in industrial applications extends to any robotics and automation application which requires learning from complex demonstrations containing series of subtasks.
\end{abstract}

% AAMAS: the ACM CCS are encouraged but optional within AAMAS papers
%%
%% The code below should be generated by the tool at
%% http://dl.acm.org/ccs.cfm
%\begin{CCSXML}
%<ccs2012>
%  <concept_id>10010520.10010553.10010562</concept_id>
%  <concept_desc>Computer systems organization~Embedded systems</concept_desc>
%  <concept_significance>500</concept_significance>
%  <concept_id>10010520.10010575.10010755</concept_id>
%  <concept_desc>Computer systems organization~Redundancy</concept_desc>
%  <concept_significance>300</concept_significance>
%  <concept_id>10010520.10010553.10010554</concept_id>
%  <concept_desc>Computer systems organization~Robotics</concept_desc>
%  <concept_significance>100</concept_significance>
%  <concept_id>10003033.10003083.10003095</concept_id>
%  <concept_desc>Networks~Network reliability</concept_desc>
%  <concept_significance>100</concept_significance>
%</ccs2012>  
%\end{CCSXML}
%
%\ccsdesc[500]{Computer systems organization~Embedded systems}

  % put your semicolon-separated keywords here!

\maketitle

%%%%%%%%%%%%%%%%%%%%%%%%%%%%%%%%%%%%%%%%%%%%%%%%%%%%%%%%%%%%%%%%%%%%%%%%%%%%%%%%%%%%%%%%%%%%%%%%%%%%%%%%%
%% start of main body of paper

\section{Introduction}

The requirement of a predefined reward function limits the practical application of Reinforcement Learning (RL) methodologies to complex problems. 
Hand engineering a reward function in a fully observable environments such as arcade style computer games \cite{brockman2016openai,kempka2016vizdoom} is possible, yet becomes far more difficult for complex 
tasks such as autonomous driving, where we have to consider multiple factors (i.e safety, fuel economy, travel time, etc ) that affect the reward signal.

To counter this problem researchers widely utilise imitation learning procedures \cite{schaal1999imitation,stadie2017third,ratliff2009learning}, where the model learns the policy following expert demonstrations,
instead of directly learning the policy through the reward signal. Most recently a Generative Adversarial Imitation Learning (GAIL) approach was proposed in \cite{ho2016generative} and further 
augmented in \cite{li2017inferring} to automatically discover the latent factors influencing human decision making. These approaches offer greater flexibility as they result in a model-free imitation learning platform; however, we observe that existing GAIL approaches generate short term responses instead of attaining long term planning. 

This work propose a novel data driven model for executing complex tasks that require long term planning. Instead of deriving a policy while only observing the current state \cite{ kuefler2017imitating, ho2016generative}, we model the state temporally with the aid of external memory modules. This results in the proposed approach having both the ability to discriminate between the sub level tasks, and determine their sequential relationships to successfully complete an overall task; as the proposed method automatically determines the optimal action to perform in the current temporal context.
 
In a typical real world environment, a task such as driving involves a series of sub tasks including, turn, overtake, follow a lane, merge into a free way, \ldots, etc.
Hence a standard demonstration from a human expert would include a mixture of such sub tasks and each would have different initial states. Even though \cite{li2017inferring, kuefler2017imitating, ho2016generative} are highly effective at discriminating between sub-tasks, 
we observe that these methods fail to identify the temporal context in which the task was performed due to inheritant architectural deficiencies. The neural network architectures proposed in \cite{kuefler2017imitating, ho2016generative} simply map the current state to an action without considering the temporal context (i.e. what has happened previously), which may impact the interpretation of the current state. Understanding temporal context is essential to interpreting how the different sub tasks are sequentially linked to each other. As such at simulation time those methods \cite{li2017inferring, kuefler2017imitating, ho2016generative} cannot decide on the optimal action to perform in the current environmental setting in order to successfully compete the overall task, as they do not have the capacity to oversee the task in it's entirety, owing to the problem being presented a single state embedding at a time. 

Furthermore, depending on the skill of the expert demonstrator, social values and individual likes and dislikes, different experts may vary their behaviour when presented similar contexts. For instance in the same autonomous driving setting, a certain expert may decide to give way to the merging traffic where as another driver may not and show aggressive behaviour. We show that these different strategies of different experts for similar context can be disentangled by attending over historical states of the expert demonstration, and learning different ways to behave under similar conditions. 

%We propose a novel data driven method for learning complex strategies automatically. Instead of deriving a policy while only observing the current state \cite{ kuefler2017imitating, ho2016generative}, we model it temporally with the aid of external memory modules. Not only it has the capability to discriminate between the sub level tasks, it also has the ability to determine their sequential relationships and successfully complete the overall task, as the proposed method automatically determine the optimal action to perform at the current temporal context. 

We incorporate two separate memories, namely global and local memories, in order to capture these two tiers of the context. The local memory is used to model the expert behaviour during each demonstration, and is reset at the end of the demonstration. In contrast, the global memory maps expert preferences during a batch of demonstrations. Local memory allows us to attend over different sub tasks within a particular trajectory and learn how these sub level tasks are sequentially linked to each other. We attend over the global memory to understand how different experts vary their behaviour under similar context and interpret how these different choices render different policies. 

Our experimental evaluations with applications to autonomous driving demonstrate the ability of the proposed model to learn the semantic correspondences in human decision making using complex expert demonstrations. The novel contributions of this paper are summarised as follows: 
\begin{itemize}
\item We introduce a novel architecture for GAIL which captures the sequential relationships in human decision making encoded within complex expert demonstrations. 
\item We incorporate neural memory networks into the imitation learning problem where it learns to automatically store and retrieve important facts for decision making without any supervision.
\item We utilise two memory modules, one for capturing sub task level dependencies and one for modelling social and behavioural factors encoded within diverse expert demonstrations.
\item We demonstrate how the hidden state representation of the memory can be utilised as a powerful information queue to augment the reward signal and generate smooth state transitions between the sub tasks.
\item We provide extensive evaluations of the proposed method with applications to autonomous driving using the TORCS \cite{wymann2000torcs} driving simulator, where the proposed method is capable of learning different expert policies and outperforms state-of-the-art methods. 
\end{itemize}

\section{Preliminaries}
An infinite horizon, discounted Markov decision process (MDP) can be represented as a tuple ($\mathcal{S}, \mathcal{A}, P, r, \rho0, \Upsilon$), where $\mathcal{S}$ represents the state space, $\mathcal{A}$ represents the action space, $P: \mathcal{S} \times \mathcal{A}  \times \mathcal{S} \rightarrow \mathbb{R}$ denotes the transition probability distribution, $r:  \mathcal{S} \rightarrow \mathbb{R} $ denotes the reward function, $\rho0:  \mathcal{S} \rightarrow \mathbb{R} $ is the distribution of the initial state, and $\Upsilon \in (0, 1)$ is the discount factor. We denote a stochastic policy $\pi: \mathcal{S} \times \mathcal{A} \rightarrow [0, 1]$, and $\pi_E$ denotes the expert policy. 
An expert demonstration $\tau_{E}$ of length $\zeta$ is generated following $\pi_E$ and is composed of state ($s_t$) and action ($a_t$) pairs, 
\begin{equation}
\tau_{E}=[(s_1,a_1), (s_2,a_2), \ldots, (s_\zeta,a_\zeta)].
\end{equation}
Then, the traditional GAIL \cite{ho2016generative} objective can be denoted as,
\begin{equation}
\mathrm{min}_\theta\mathrm{max}_wV(\theta,w)=\mathbb{E}_{\pi_{\theta}} [\mathrm{log} D_{w}(s,a)] + \mathbb{E}_{\pi_{E}} [\mathrm{log} D_{w}(s,a)],
\label{eq:traditional_GAIL}
\end{equation}
where policy $\pi_{\theta}$ is a neural network parameterised by $\theta$ which generates the policy imitating $\pi_{E}$, and $D_w$ is the discriminator network parameterised by $w$ which tries to distinguish state-action paris from $\pi_{\theta}$ and $\pi_E$. $\mathbb{E}_{\pi}[f(s,a)]$ denotes the expectation of $f$ over state action pairs generated by policy $\pi$.

In \cite{li2017inferring} the authors emphasise the advantage of using Wasserstein GANs (WGAN) \cite{arjovsky2017wasserstein} over the traditional GAIL objective as this method is less prone to vanishing gradient and mode collapse problems. The WGAN objective transferred to GAIL can be written as,
\begin{equation}
\mathrm{min}_\theta\mathrm{max}_wV^*(\theta,w)=\mathbb{E}_{\pi_{\theta}} [ D_{w}(s,a)] + \mathbb{E}_{\pi_{E}} [ D_{w}(s,a)].
\label{eq:w_GAIL}
\end{equation}
In this approach the discriminator assigns scores to to its inputs, trying to maximise the score values for the expert policy $\pi_E$ while minimising the score for generated policy $\pi_{\theta}$, in contrast to Eq. \ref{eq:traditional_GAIL} which tries to classify the two policies. 

%In the event of expert trajectories coming from mixture of experts, with different skill levels, habits and strategies, authors in \cite{} extend the GAIL method to recover both policy and the latent structures influencing those policies. They denote the expert policy $\pi_E$ as $\pi_E(a|s,c)$ where $c$ is a latent variable parameterising the structure of expert decision making and $c \sim p(c)$ which is the prior distribution of $c$.
%
%This requires maximisation of mutual information between $c$ and the state-action pairs which can be denoted by $I(c; s,a)$. The authors in \cite{} shows how this can be incorporated into Eq. \ref{eq:w_GAIL} by defining lower bound to $I(c; s,a)$ as,
%\begin{equation}
%L_1(\pi_{\theta},Q)=\mathbb{E}_{c \sim p(c), a \sim \pi_{\theta}(. | s, c) }[log Q(c |s,a)],
%\end{equation}
%where $Q(c |s,a)$ us an approximation of $P(c |s,a)$ parameterised by $\psi$. Then Eq. \ref{eq:w_GAIL} under this regularisation can be denoted as,
%\begin{equation}
%\mathrm{min}_{\theta,\psi}\mathrm{max}_wV^*(\theta,w)-\lambda_1L_1(\pi_{\theta},Q_\psi),
%\label{eq:info_GAIL}
%\end{equation}
%where $\lambda_1 > 0$ is a hyper parameter. Yet, as show in \cite{}, the extracted demonstrations from complex environment would typically contain a series of sub tasks in different order and different initial states making distinguishing different policies simply considering state-action paris not optimal. 

\section{Related Work}
Imitation learning focuses on building autonomous agents that can acquire skills and knowledge from the observed demonstrations of experts  \cite{schaal1999imitation, calinon2009robot, argall2009survey}. 

There exists two major lines of work for imitation learning: Behavioural Cloning (BC) \cite{pomerleau1989alvinn, ross2011reduction} which performs direct supervised learning to match observations to actions; and Inverse Reinforcement Learning (IRL) \cite{ng2000algorithms, finn2016guided, abbeel2004apprenticeship} which tries to recover the reward function followed by the experts assuming that the experts follow an optimal policy with respect to the reward function. 

Even with the recent advances in deep learning techniques, BC approaches tend to suffer from the problem of cascading errors \cite{ross2010efficient} during simulations \cite{morton2017simultaneous,kuefler2017imitating}. For instance, BC based autonomous driving methods \cite{Fernando_2017_ICCV, chen2015deepdriving, hadsell2009learning, aly2008real} rely heavily on larger training databases in order to generalise for different states, yet fail in practise due to small inaccuracies compounding sequentially \cite{kuefler2017imitating}. 

In contrast IRL methods generalise more effectively \cite{finn2016connection} and have been applied multiple times for modelling human driving behaviour \cite{gonzalez2016high,sadigh2016planning}. 

Yet learning with IRL is computationally expensive as one should first recover the reward function and then apply RL techniques with that reward function in order to find the policy. Hence a third line of work has emerged, namely GAIL, which attempts to imitate the expert behaviour through direct policy optimisation rather than learning the reward function first. 

Some recent studies \cite{li2017inferring,kuefler2017imitating} have considered the problem of modelling human highway driving behaviour at a simpler sub task level, but haven't considered learning to perform complex tasks which are composed of a mixture of sequentially linked sub tasks. 

We draw our inspiration from \cite{duan2017one,our_wacv1,fernando2017tree,parisotto2017neural} where an external memory module is used to capture the current context of the autonomous agent. However these memory architectures are problem specific. For instance in \cite{parisotto2017neural} the authors design a memory module to store a map of explored areas for a 3D navigation task where as \cite{duan2017one} uses a memory module to store coordinates of the blocks in robot block stacking experiments. Consequently, the structure and operations of the memory modules are specifically tailored for those tasks.

Distinctively, we propose a generalised framework for GAIL which uses memory modules to store local and global salient information and automatically learn the temporal relationships. 

Similar to \cite{duan2017one, parisotto2017neural} we also rely on a soft attention mechanism \cite{bahdanau2014neural} to compare the current state representation with the content of the memories, which is widely a used mechanism to learn such dependencies in a variety of problem domains, including human navigation \cite{fernando2017soft+,our_wacv2}, image captioning \cite{xu2015show}, neural machine translation \cite{kumar2016ask}.  
 
\section{Architecture}
\begin{figure*}
\includegraphics[width=.85\linewidth]{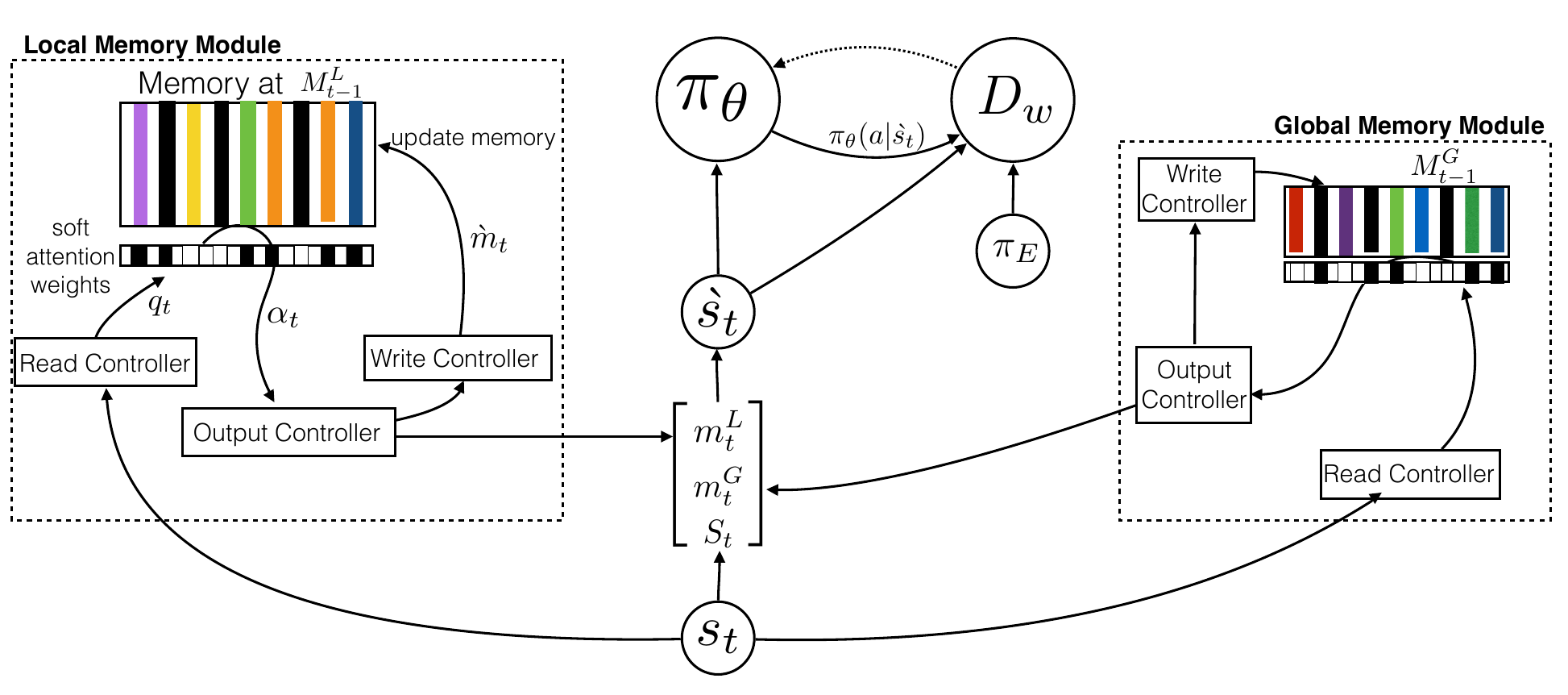}
\caption{The proposed MA-GAIL model: The model is composed of a policy generator $\pi_\theta$, a discriminator $D_w$ and two memory modules for capturing local $M^L$ and global $M^G$ contexts. The input state embedding $s_t$ triggers a read operation in both memories which generates a query $q_t$ to question the temporal relationships between the current state and the content of the memories, $M^L_{t-1}$ and $M^G_{t-1}$. Then using soft attention, we generate a weighted distribution $\alpha_t$ quantifying the similarity between each memory slot and the query vector. This is used by the output controller to generate a memory output and produces $m_t^L$ and $m_t^G$ from the respective memory modules. The write control module of each memory updates the respective memories accordingly. We directly concatenate the $m_t^L$ and $m_t^G$ vector representations together with state embedding $s_t$ to generate an augmented state representation $\grave{s_t}$. The policy generator accepts this as the input and outputs an action for that particular state. The discriminator also uses the augmented state in order to provide feedback to guide the policy generator using the expert policy $pi_E$ as the guideline.}
\label{fig:architecture}
	\vspace{-2mm}
\end{figure*}

This section describes our approach to disentangle the mixture of expert policies at the task level and learn the different strategies employed by different experts in order to complete various sub goals. The high level architecture of the proposed Memory Augmented Generative Adversarial Imitation Learning (MA-GAIL) network is shown in Fig. \ref{fig:architecture}. In contrast to traditional GAIL \cite{kuefler2017imitating, ho2016generative} systems, which utilise only a discriminator ($D_w$) and a policy generator $\pi_\theta$, we additionally use two memory modules, named local memory $M^L$ and global memory $M^G$.

Memory stores important facts from the expert trajectories and attends to them systematically to retrieve facts relevant to the current state. For instance in autonomous driving, visual scene information provides powerful cues and heavily contributes to decision making. Hence if our memory is composed of scene embeddings, by attending over the relevant memory embeddings with similar scene dynamics to the current state, one can retrieve important clues to aid decision making. 

Furthermore, as outlined in previous sections, expert decision making in similar situations may vary significantly due to numerous social and behavioural factors. This inspires the need for a global memory to capture, compare and contrast the different expert behaviours under similar contexts and learn different ways to behave. 

In the following subsection we describe the neural memory architecture that we use for global and local memory modules and later in Sec. \ref{sec:mgail} we explain how we incorporate those memory modules with the GAIL objective. 

\subsection{Neural Memory Module}
\label{sec:neural_memory}
As shown in Fig. \ref{fig:architecture} each memory module is composed of 1) a memory stack to store the relevant data; 2) a read controller to query content from the memory stack; 3) a write controller to compose and update the memory; and 4) an output controller to send through the output of the read operation. 

Formally, let $M \in  \mathbb{R}^{k \times l}$ be the memory stack with $k$ memory slots, and $l$ be the embedding dimension of the state representation $s$. The representation of the memory at time instance $t-1$ is given by $M_{t-1}$. First the read controller uses a read function to generate a query vector ${q_t}$ such that,
\begin{equation}
 q_t=f^{LSTM}_r(s_t),
\label{eq:mem_read}
\end{equation}
where $f^{LSTM}_r$ is a read function which uses Long Short Term Memory (LSTM) \cite{hochreiter1997long} cells and $s_t$ is the current state of the episode. 

Then we generate a score vector $a_t$ over each memory slot representing the similarity between the current memory state $M_{t-1}$ and the generated query vector ${q}_t$ by attending over the memory slots such that,
\begin{equation}
 a_{(t, j)}={q}_tM_{(t-1, j)},
\label{eq:attention}
\end{equation}
where $M_{(t-1, j)}$ denotes the content of the $j^{th}$ memory slot of the memory at $t-1$ where $j=[1, \ldots, k]$. Then the score values are normalised using soft attention \cite{bahdanau2014neural}, generating a probability distribution over each memory slot as follows,

\begin{equation}
 \alpha_{(t,j)}=\dfrac{\mathbb{E}(a_{(t, j)})}{ \sum_{j=1}^{k} \mathbb{E}(a_{(t, j)})}.
\label{eq:h_t}
\end{equation}
Now the output controller can retrieve the memory output for the current state by,
\begin{equation}
 m_t=\sum_{j=1}^{k} \alpha_{(t,j)}M_{(t-1, j)}.
\label{eq:mem_out}
\end{equation}
Then we generate a vector $\grave{m}_t$ for the memory update by passing the output of the memory through a write function $f_w^{LSTM}$ composed of LSTM memory cells, 
\begin{equation}
 \grave{m}_t=f^{LSTM}_w(m_t),
\label{eq:mem_write}
\end{equation}
and update the memory using,
\begin{equation}
 M_t=M_{t-1}(I-\alpha_t \otimes e_k)^T + (\grave{m}_t \otimes e_l)(\alpha_t \otimes e_k)^T,
\label{eq:mem_write}
\end{equation}

where $I$ is a matrix of ones, $e_l \in  \mathbb{R}^l $ and $e_k \in  \mathbb{R}^k$ are vectors of ones and $\otimes$ denotes the outer product which duplicates its left vector $l$ or $k$ times to form a matrix. 

The functions $f^{LSTM}_r$ and $f^{LSTM}_w$ use the gated operations defined in \cite{hochreiter1997long} and shown below in Equations \ref{eq:lstm_forget_gate} to \ref{eq:lstm_hidden_state}. Instead of simply generating a query to read, or an update vector to hard write the current memory content, LSTM based read and write operations allow us to retain long term histories of those operations and decide upon how much the query or memory update vector should differ based on the historical read/ write operations. We define the operations of $f^{LSTM}_r$ and $f^{LSTM}_r$ as,
\begin{equation}
f_t =\sigma(w_f[h_{t-1}, x_t]), 
\label{eq:lstm_forget_gate}
\end{equation}
\begin{equation}
i_t = \sigma(w_i[h_{t-1}, x_t]), 
\end{equation}
\begin{equation}
\grave{c}_t = \mathrm{tanh}(w_c[h_{t-1}, x_t]),
\end{equation}
\begin{equation}
c_t = f_t \cdot c_{t-1} + i_t \cdot \grave{c}_t ,
\end{equation}
\begin{equation}
o_t = \sigma(w_o[h_{t-1}, x_t]),
\label{eq:lstm_output}
\end{equation}
\begin{equation}
h_t = o_t \cdot \mathrm{tanh}(c_t),
\label{eq:lstm_hidden_state}
\end{equation}

where $\sigma(\cdot)$ is the sigmoid activation function and $w_f, w_i, w_c, w_o$ are the weight vectors for forget, input, cell state and output gates of the LSTM cell. $h_{t-1}$ denotes the hidden state of the LSTM cell at time instance $t-1$ where as $c_{t-1}$ denotes the cell state of the LSTM cell at the same time instance. The input to the cell at time $t$ is given as $x_t$. When applied these equations to the proposed approach, for Eq. \ref{eq:mem_read} the LSTM input is the current state $s_t$, and for Eq. \ref{eq:mem_write} it is the memory output vector $m_t$. The LSTM cell output is denoted $o_t$ in Eq. \ref{eq:lstm_output} and is replaced with $q_t$ and $\grave{m}_t$ for Eq. \ref{eq:mem_read} and Eq. \ref{eq:mem_write} respectively.

\subsection{Memory Augmented Generative Adversarial Imitation Learning (MA-GAIL)}
\label{sec:mgail}

We deploy two instances of the neural memory module introduced in Sec. \ref{sec:neural_memory}, functioning as local and global memories denoted as $M^L_t$ and $M_t^G$ respectively in Fig. \ref{fig:architecture}. Note that via passing the state embeddings to the memory, our autonomous agent can question the current temporal context that it is in. 

For the local memory, as the state embeddings for similar sub tasks would have similar features, they would generate stronger activations allowing the model to estimate the sub task that it is currently in. 

In contrast, global memory contains a batch of expert trajectories. Hence the memory output $m_t^G$ of the global memory would have distributions for different ways of how an expert would react in the current state, capturing their different strategies. 

In the proposed approach we directly concatenate the physical state of the system $s_t$ with the relevant memory outputs, $m_t^L$ and $m_t^G$, from local and global memories respectively, to produce an augmented state,
\begin{displaymath}
 \grave{s}_t=\Bigg[
  \begin{tabular}{c}
  $s_t$ \\
  $m_t^L$  \\
  $m_t^G$ 
  \end{tabular}
 \Bigg] .
\end{displaymath}

Training a policy $\pi_{\theta}(a | \grave{s}_t, c)$ on the augmented state representation generates a dynamic policy which can fully utilise memory states to capture expert dynamics. We would like to note that we do not provide any guidance on what memory content should be extracted or updated. The read/ write controllers on each memory module learn a distribution over the input states and learn salient factors that they should focus on in order to maximise the overall GAIL performance. 

Now we modify the GAIL objective in Eq. \ref{eq:w_GAIL} with the augmented state to obtain the proposed MA-GAIL,
\begin{equation}
\mathrm{min}_{\theta}\mathrm{max}_w\grave{V} =\mathrm{min}_{\theta,w}\mathrm{max}_w\mathbb{E}_{\pi_\theta}[D_w(\grave{s},a)] - \mathbb{E}_{\pi_E}[D_w(\grave{s},a)],
\label{eq:MA_GAIL}
\end{equation}
with $\pi_\theta$ updated by the TRPO method introduced in \cite{schaul2015prioritized} and the discriminator $D_w$ updated with RMSprop \cite{tieleman2012lecture}.

\subsection{Reward augmentation with memory states}
\label{sec:r_a}
When designing autonomous agents it is desirable to have smooth transitions between the sub tasks. For example in an autonomous driving setting, when exiting the freeway from an exit ramp, the agent should learn to smoothly shift between the sub tasks such as  merge, slow down and follow the exit ramp; all the while avoiding sudden braking, accelerations and  turns. %Yet in models such as \cite{li2017inferring, kuefler2017imitating, ho2016generative} we observe undesirable behaviour such as oscillation in acceleration and breaking, when the agent shifts between different modes. 

A naive way to solve this problem is to penalise sudden action changes, but this limits the capability of the agent. For instance by limiting steering angels between consecutive actions we eradicate the capability of the agent to perform u-turns, or by controlling the acceleration we might reduce the capability of driving at high speed on a freeway. Similar problems exist if we penalise sudden state transitions. For the same autonomous driving example, where the state is composed of visual inputs of the road scene, scene changes such as illumination variations, merging traffic or changes in road conditions result in diverse changes in the state embeddings. Hence considering state embeddings alone leads to erroneous behaviour. 

We overcome this problem by observing the changes in local memory states. The local memory generates temporal attention over a series of state embeddings, rectifying sudden fluctuations in state due to the various local factors listed above. Hence changes in local memory attention occur only due to the sub task changing as it captures salient aspects of the trajectory, and shifts in attention denoting the changes in the behaviour. 

We monitor the dispersion between local memory outputs in consecutive time steps and penalise sudden changes. Let $m_{t-1}^L$ and $m_t^L$ be the respective local memory outputs for input states $s_{t-1}$ and $s_t$. We define a function $f^*$ which extracts memory outputs following Eq. \ref{eq:mem_read} to Eq. \ref{eq:mem_out},
\begin{equation}
m^L_t=f^*(s_t)
\end{equation}

Now we can define a penalty for memory state change using the cosine distance \cite{nguyen2010cosine} between the consecutive memory states as, 
\begin{equation}
\eta(\pi_\theta)= \mathbb{E}_{s_t,s_{t-1} \sim \pi_\theta}[cos(m^L_t- m^L_{t-1})],
\end{equation}
where $cos(m^L_t- m^L_{t-1})$ is given by,
\begin{equation}
\begin{split}
cos(m^L_t- m^L_{t-1}) &=1-\cfrac{m^L_t \cdot m^L_{t-1}}{||m^L_t||_2 ||m^L_{t-1}||_2}
\\
 & =1- \cfrac{\sum\limits_{j=1}^{l}{m^L_{(t,j)}}{m^L_{(t-1,j)}}}{\sqrt{\sum\limits_{j=1}^{l}({m^L_{(t,j)}} )^2}\sqrt{\sum\limits_{j=1}^{l}({m^L_{(t-1,j)}})^2}} ,
\end{split}
\end{equation}

where $m^L_{(t,j)}$ and $m^L_{(t-1,j)}$ are the components of the memory vectors $m^L_t$ and $m^L_{t-1}$ and $j=[1, \ldots, l]$ where $l$ is the embedding dimension of the state. Then the combined MA-GAIL objective with a reinforced objective for smoothed state transitions can be denoted as,
\begin{equation}
\mathrm{min}_{\theta}\mathrm{max}_w\grave{V} + \lambda_0\eta(\pi_\theta),
\label{eq:MA_GAIL2}
\end{equation}

where $\lambda_0 > 0$ is a hyper parameter which controls the tradeoff between imitation and reinforcement learning objectives. 

\section{Evaluations and discussion}

In many real world applications, representing the state, $s$, in relation to visual inputs such as images or image sequences is highly beneficial. It is often inexpensive to obtain and highly informative \cite{li2017inferring,Fernando_2017_ICCV}. Furthermore, direct learning from visual inputs, rather than hand crafting the state features, enables the policy generator to be directly transferable from one domain to another. Hence we demonstrate the performance of the proposed method with applications to autonomous driving from visual inputs. 

\subsection{Environment setup}
In \cite{li2017inferring} the authors provide an API for The Open Racing Car Simulator (TORCS) \cite{wymann2000torcs} similar to OpenAI Gym \cite{brockman2016openai}, which generates a realistic dashboard view and driving related information. 

We collected human expert demonstrations by manually driving the vehicle along the racing track. We could not use the demonstrations provided in \cite{li2017inferring} as each of those episodes contains only a single sub level task such as overtaking and turning. We believe it is an over simplification of the real world driving task where a typical demonstration would contain a series of such sub level tasks. 

Therefore we collected 150 expert trajectories from 3 experts with each demonstration lasting 500 frames. Within each expert trajectory a series of driving tasks such as lane change, turns, staying within a lane, and overtaking are included. 

\subsection{Network structure}

\begin{figure}
\subfigure[State embedding]{\includegraphics[width=.95\linewidth]{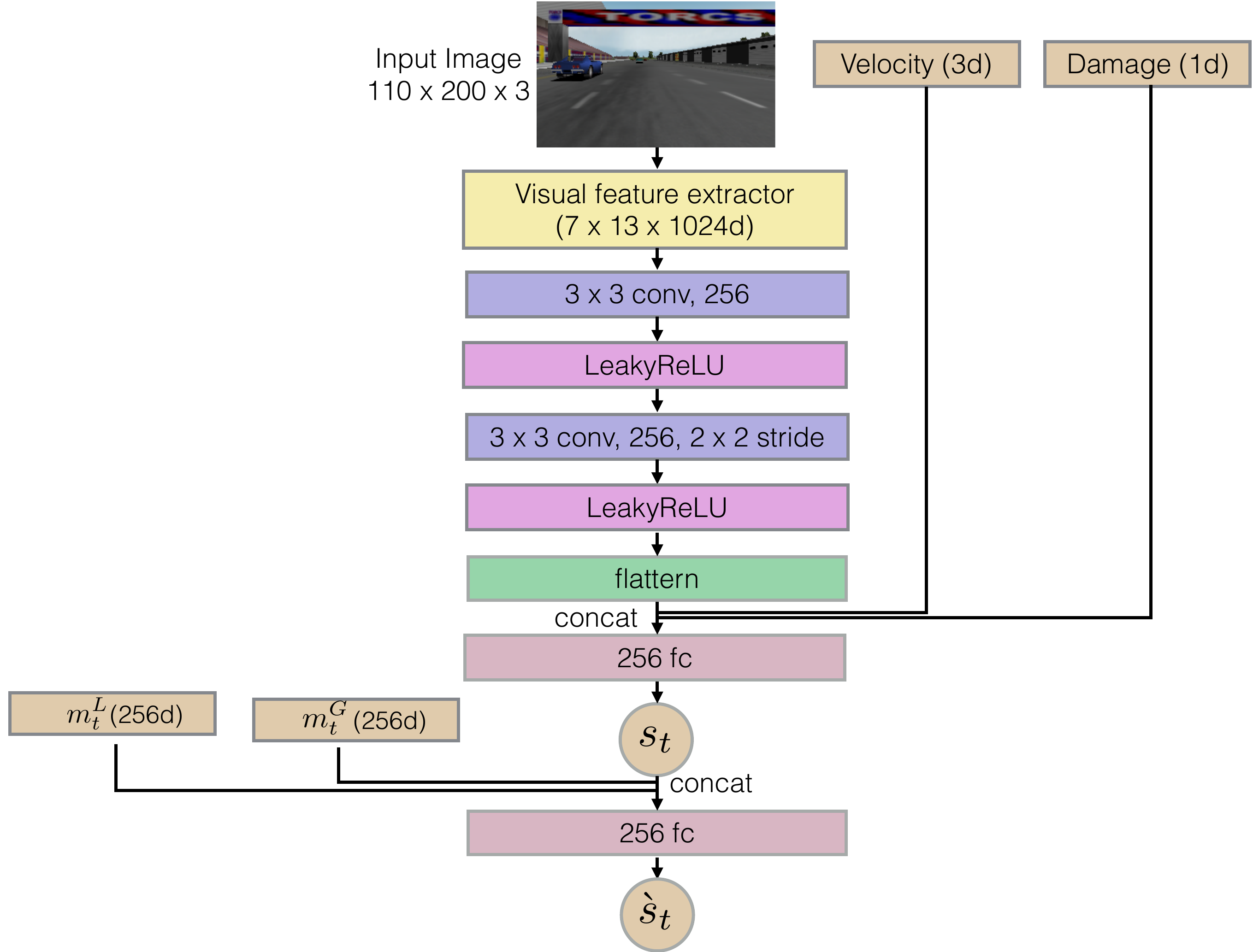}}
\subfigure[Policy generator $\pi_\theta$]{\includegraphics[width = .370 \linewidth]{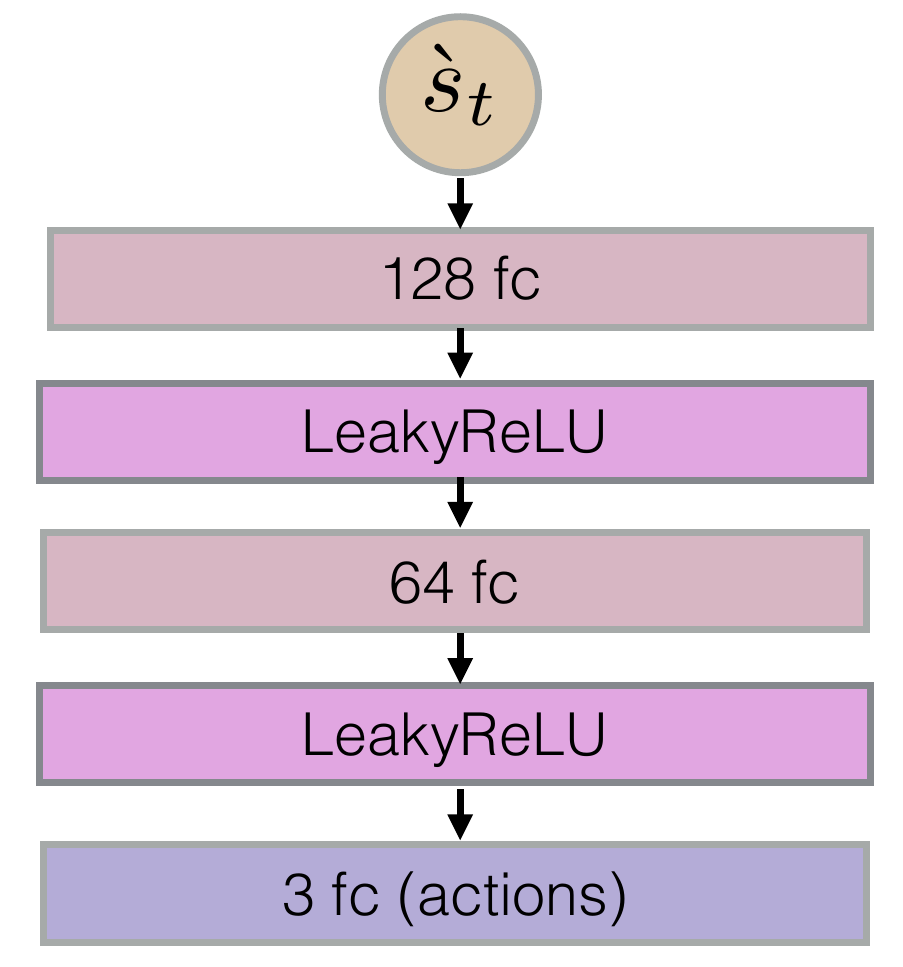}}
\subfigure[Discriminator $D_w$]{\includegraphics[width = .480 \linewidth, height = .390 \linewidth]{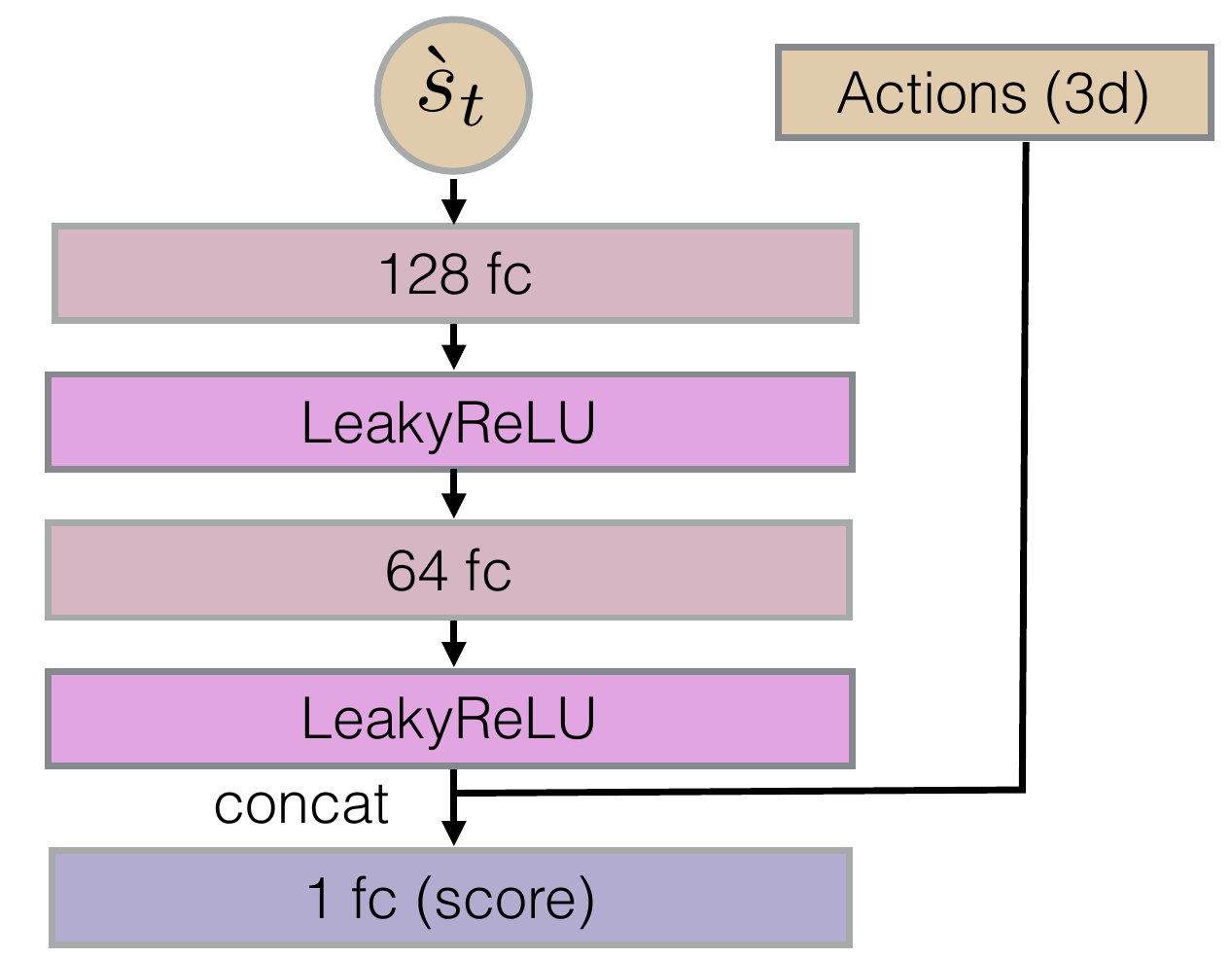}}
\caption{Network architecture of the proposed MA-GAIL model. $conv$ denotes a convolution layer, and $fc$ denotes a fully connected layer. (a) State embedding: The input image is passed through a pre-trained visual feature extractor and later combined with velocity and current damage of the vehicle to generate a current state representation. We combine this with memory outputs $m_t^L$ and $m_t^G$ of local and global memories to generate the augmented state embedding. (b) Policy generator: accepts the augmented state embedding as the input and generates a 3 dimensional action to take at that particular state. (c) Discriminator: Uses the augmented state embedding and the action to output a score value to discriminate the policy that is generated by the input state-action pair. It gives a higher score value to expert policies $\pi_E$ and a lower score to $\pi_\theta$. }
\label{fig:model}
	\vspace{-2mm}
\end{figure}

The TORCS environment outputs a dashboard camera view, the current speed in x, y and z dimensions of the car, and a real value representing the damage of the car. Fig \ref{fig:model} illustrates our approach to combine these diverse information sources to a single state representation.

We use the Keras \cite{chollet2017keras} implementation of the Deep Residual Network (DRN) \cite{he2016deep} pre-trained on the ImageNet classification task \cite{deng2009imagenet}, as our pre-trained feature extractor to extract salient information from the visual input. We pass each input frame through the DRN and extract the activations from the $40^{th}$ activation layer. Then we apply a series of convolution operations to compress our spatial feature representation and construct a 256 dimension representation of the current state $s_t$. 

We combine this representation with local $m^L_t$ and global $m_t^G$ memory outputs and generate an augmented state $\grave{s_t}$ by passing those embeddings through a series of fully connected layers. 

Note that each memory stack $M^L$ and $M^G$, is of dimension $l \times k$ where $l=256$ is the embedding dimension of the state $s_t$, and $k$ is the number of memory slots. For local memory we set $k=500$ because each expert trajectory is composed of 500 frames. Global memory should retain information for a batch of expert demonstrations, hence $k=500 \times 50$ where 50 is the batch size. 

Our policy generator Fig \ref{fig:model} (b) uses this augmented state representation as its input and outputs a three dimensional action representing [\textit{steering}, \textit{acceleration}, \textit{breaking}].

The discriminator $D_w$, shown in Fig. \ref{fig:model} (c) also accepts the augmented state embedding ($\grave{s}_t$) along with the current action $a_t$ and outputs a score value for the policy that generated input state action pair. As per \cite{li2017inferring} we first train the model with behavioural cloning using random weights, and the learned weights initialise the network for imitation learning

\subsection{Evaluation of driving behaviour}

\subsubsection{Baseline models}

We compare our model against 5 state-of-the-art baselines. The first baseline model we consider is a Static Gaussian (SG) model which uses an static gaussian distribution $\pi(G|s) = N(a | \mu, \sigma)$ which assumes an unchanged policy through out the simulation, and is fitted using maximum likelihood \cite{vroman2014maximum}. The second baseline is a Mixture Regression (MR) \cite{lefevre2014comparison} model which is a gaussian mixture over the joint space of the actions and state features and is trained using expectation maximisation \cite{friedman2001elements}. The third baseline we compare our model against is the GAIL-GRU model proposed in \cite{kuefler2017imitating} which optimises a recurrent policy. The first 3 baselines only accept a hand crafted feature representation of the state. Hence, following a similar approach to  \cite{kuefler2017imitating} we extract the features listed in Tab. \ref{tab:features} and used the implementations provided by the authors of \cite{kuefler2017imitating} and available online \footnote{https://github.com/sisl/gail-driver}.
Our next baseline is the Info-GAIL model presented in \cite{li2017inferring} and uses a visual feature representation of the scene along with the auxiliary information (i.e velocity at time $t$, previous actions at time $t-1$ and $t-2$, damage of the car) to represent the current state. In order to emphasise the importance of the imitation learning strategy rather than simply cloning the behaviour of the expert using behavioural cloning, we also provide evaluations against the Behavioural Cloning (BC) method given in \cite{li2017inferring}. For the Info-GAIL model and BC model we use the implementation released by the authors \footnote{https://github.com/YunzhuLi/InfoGAIL}.

\begin{table*}
  \caption{Handcrafted feature for the SG, MR and GAIL-GRU baselines}
  \label{tab:features}
  \begin{tabular}{ccl}
    \toprule
    Feature & Range &Description \\
    \midrule
    Angle & [-180 180] & Angle between the car direction and the direction of the track axis\\
    Track & [0, 200] meters & Vector of 19 range finder sensor detections, denoting the distance between the track edge and the car\\
     Track position & $ [- \infty, + \infty]$ & Distance between the car and the track axis\\
    Speed X & $[- \infty, + \infty]$ Km/h & Speed of the car along the longitudinal axis of the car\\
    Speed Y & $[- \infty, + \infty]$ Km/h & Speed of the car along the transverse axis of the car\\
    Speed Z & $[- \infty, + \infty] $Km/h & Speed of the car along the z axis of the car\\
    Wheel Spin velocity &$ [- \infty, + \infty]$ rad/s & Vector of 4 values representing the speed of each wheel of the car\\
    Front Distance & [0 ,200] meters & Distance to the closest car in front\\
    Back Distance & [0 , 50] meters & Distance to the closest cars in back\\
  \bottomrule
\end{tabular}
	\vspace{-2mm}
\end{table*}

\subsubsection{Validation}

To evaluate the relative performance of each model we simulate 500 frame length simulations 20 times each in identical environments. We also asked a human expert to drive the car in the same environment to provide a comparative upper bound on the evaluated metrics. 

It is desirable to have driver models that match real world human behaviour. Therefore we measure the dispersion between the human expert and modelled distributions over emergent quantities using Kullback-Leibler Divergence (KL) \cite{joyce2011kullback}. Similar to \cite{kuefler2017imitating}, for each model we compute empirical distributions over speed, acceleration, turn-rate, jerk, and inverse time-to-collision (iTTC) over simulated trajectories.

\begin{table}
  \caption{KL Divergence between different prediction models and human expert behaviour.} 
  \label{tab:experiment_0}
  \resizebox{.98\linewidth}{!}{
  \begin{tabular}{cccccc}
    \toprule
    Method & Speed & acceleration &  turn-rate & jerk & iTTC \\
    \midrule
    SG             & 0.43 & 0.33 & \textbf{0.24} & 1.95& 0.58\\
    MR             & 0.41 & 0.45 & 0.51 & 1.78 & 0.49 \\
    GAIL-GRU &0.38  & 1.20  & 1.90&  1.56 & 0.45\\
    BC              &0.38  & 1.40  & 1.50 & 1.24 & 0.52\\
    Info-GAIL    &0.35  & 1.00  & 0.97 & 0.91 & 0.38\\
    MA-GAIL     & \textbf{0.31} & \textbf{0.31} & 0.50 & \textbf{0.45} & \textbf{0.30}\\
    \bottomrule
  \end{tabular}
  }
	\vspace{-2mm}
\end{table}

\begin{figure*}[htb]
\subfigure[]{\includegraphics[width=.09\linewidth]{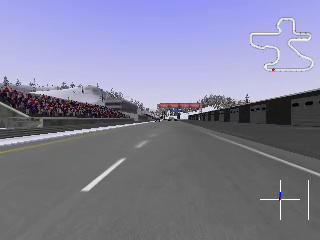}}
\subfigure[]{\includegraphics[width = .09 \linewidth]{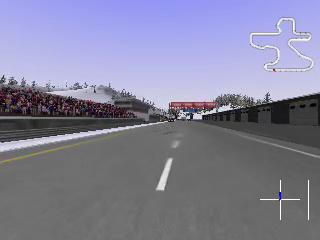}}
\subfigure[]{\includegraphics[width = .09 \linewidth]{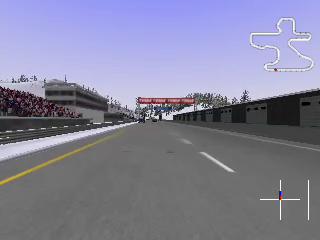}}
\subfigure[]{\includegraphics[width=.09\linewidth]{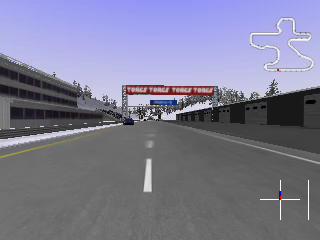}}
\subfigure[]{\includegraphics[width = .09 \linewidth]{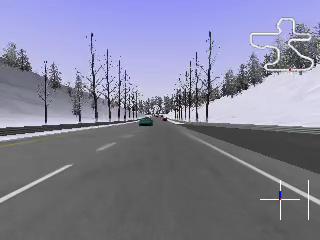}}
\subfigure[]{\includegraphics[width = .09 \linewidth]{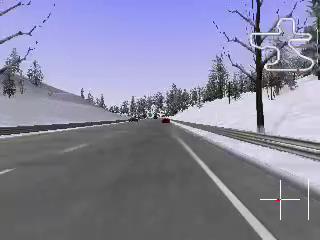}}
\subfigure[]{\includegraphics[width = .09 \linewidth]{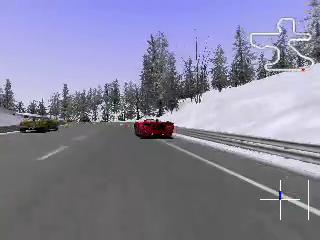}}
\subfigure[]{\includegraphics[width = .09 \linewidth]{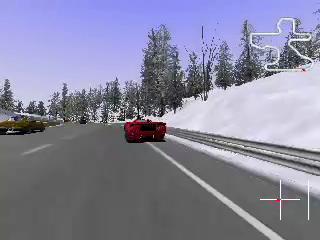}}
\subfigure[]{\includegraphics[width=.09\linewidth]{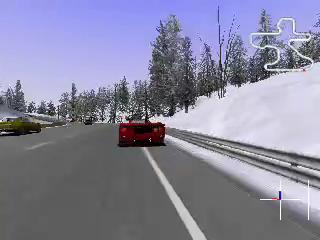}}
\subfigure[]{\includegraphics[width = .09 \linewidth]{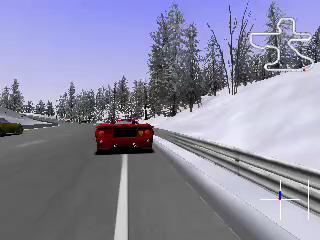}}
	\vspace{-1mm}
\subfigure{\includegraphics[width = 0.985 \linewidth]{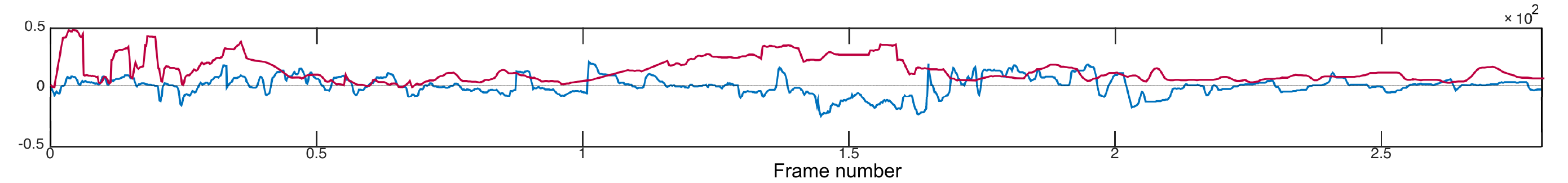}}
\caption{Visual inputs along with the predicted steering wheel angles (in blue) and acceleration (in red) for GAIL-GRU model.}
\label{fig:oscillation_GAIL}
	\vspace{-2mm}
\end{figure*}

\begin{figure*}[htb]
\subfigure[]{\includegraphics[width=.09\linewidth]{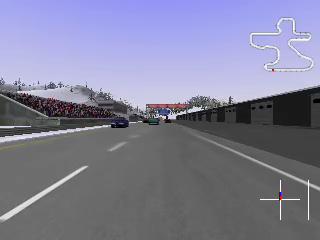}}
\subfigure[]{\includegraphics[width = .09 \linewidth]{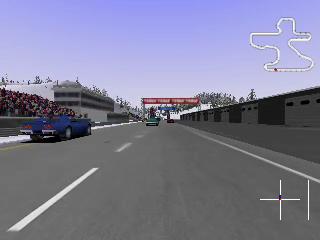}}
\subfigure[]{\includegraphics[width = .09 \linewidth]{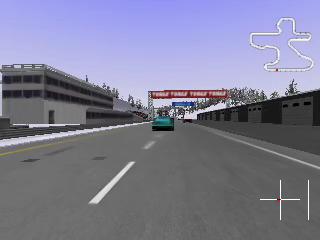}}
\subfigure[]{\includegraphics[width=.09\linewidth]{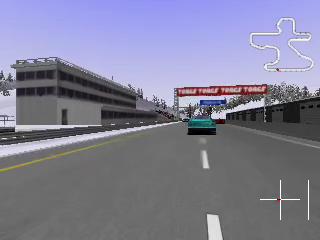}}
\subfigure[]{\includegraphics[width = .09 \linewidth]{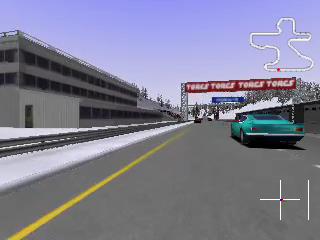}}
\subfigure[]{\includegraphics[width = .09 \linewidth]{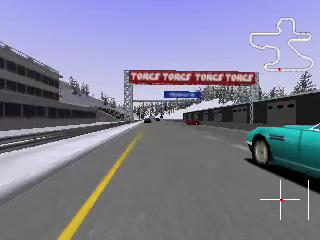}}
\subfigure[]{\includegraphics[width = .09 \linewidth]{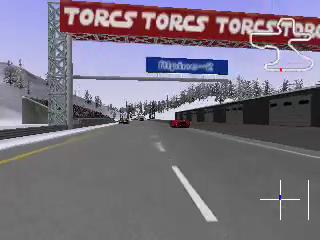}}
\subfigure[]{\includegraphics[width = .09 \linewidth]{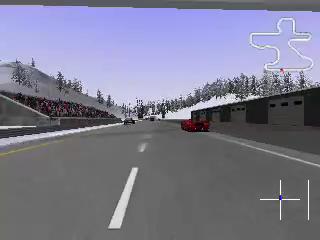}}
\subfigure[]{\includegraphics[width=.09\linewidth]{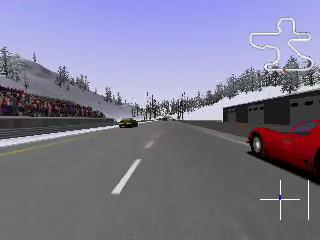}}
\subfigure[]{\includegraphics[width = .09 \linewidth]{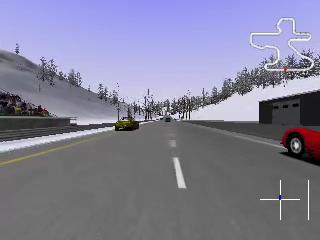}}
	\vspace{-1mm}
\subfigure{\includegraphics[width = 0.985 \linewidth]{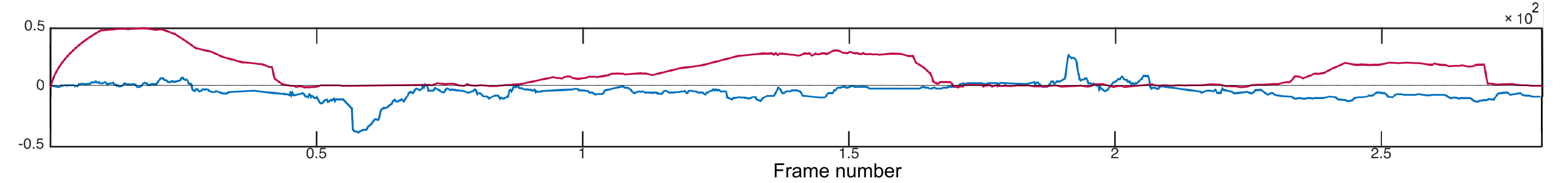}}
\caption{Visual inputs along with the predicted steering wheel angles (in blue) and acceleration (in red) for MA-GAIL model.}
\label{fig:oscillation_MA-GAIL}
	\vspace{-2mm}
\end{figure*}

\begin{figure*}[htb]
\subfigure[]{\includegraphics[width=.12\linewidth]{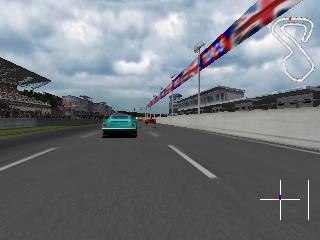}}
\subfigure[]{\includegraphics[width = .12 \linewidth]{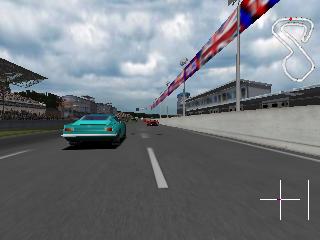}}
\subfigure[]{\includegraphics[width = .12 \linewidth]{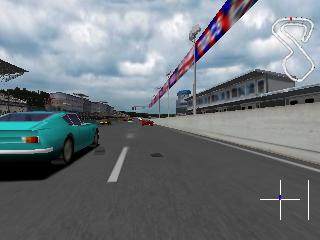}}
\subfigure[]{\includegraphics[width=.12\linewidth]{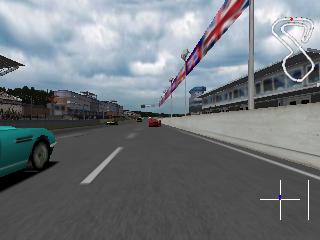}}
\subfigure[]{\includegraphics[width = .12 \linewidth]{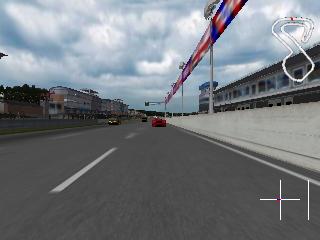}}
\subfigure[]{\includegraphics[width = .12 \linewidth]{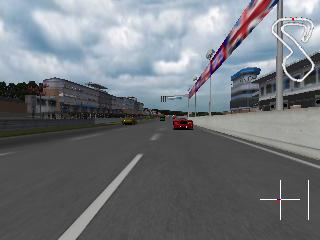}}
\subfigure[]{\includegraphics[width = .12 \linewidth]{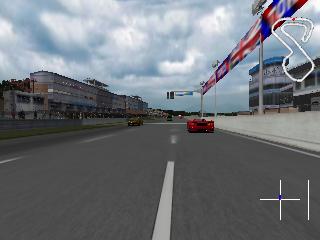}}
\subfigure[]{\includegraphics[width = .12 \linewidth]{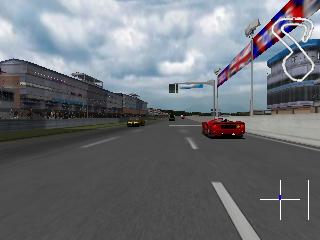}}
\subfigure[]{\includegraphics[width=.12\linewidth]{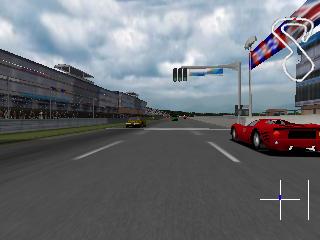}}
\subfigure[]{\includegraphics[width = .12 \linewidth]{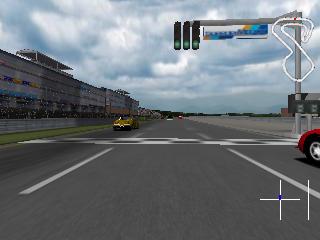}}
\subfigure[]{\includegraphics[width = .12 \linewidth]{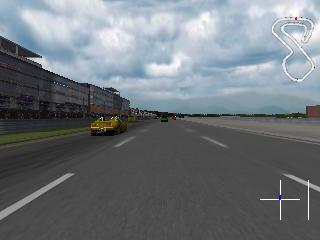}}
\subfigure[]{\includegraphics[width=.12\linewidth]{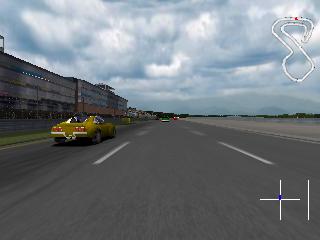}}
\subfigure[]{\includegraphics[width = .12 \linewidth]{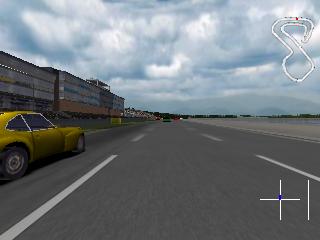}}
\subfigure[]{\includegraphics[width = .12 \linewidth]{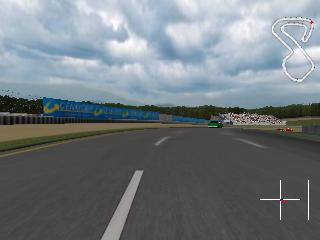}}
\subfigure[]{\includegraphics[width = .12 \linewidth]{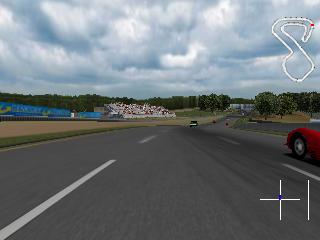}}
\subfigure[]{\includegraphics[width = .12 \linewidth]{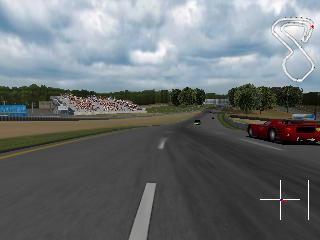}}
\caption{Visual inputs to the MA-GAIL model at different time steps of a single simulation experiment. Different sub level tasks such as lane change, over take, turn , lane following is attempted and successfully completed within a single simulation. }
\label{fig:simulation}
	\vspace{-2mm}
\end{figure*}

From the results tabulated in Tab. \ref{tab:experiment_0} we observe the poorest performance in SG model as it models a static policy. We observe that the poor steering performance of GAIL-GRU, BC and Info-GAIL is caused by oscillations of the steering angle, which can also be seen in Fig. \ref{fig:oscillation_GAIL} for GAIL-GRU. We observe that when switching between the sub tasks such as lane following, changing lanes, and overtaking the model is unsure on the present context due to the limited history it possesses and will alternate between outputting small positive and negative turn-rates rather than performing the overall task successfully (see Fig. \ref{fig:oscillation_GAIL}). This leads those models having higher dispersions in acceleration, turn-rate and jerk compared to the MA-GAIL model. With the aid of local and global memories the proposed model clearly identifies the required sub level tasks at different temporal contexts and performs them successfully, achieving better performance compared to the baselines. 

In Fig. \ref{fig:oscillation_GAIL} and \ref{fig:oscillation_MA-GAIL} we visualise the input frames along with the predicted steering wheel angles, shown in blue, and acceleration shown in red, which are quantised between $-0.5$ and $+0.5$, for GAIL-GRU and MA-GAIL models respectively. The two models observe the same initial state but it can be seen that the GAIL-GRU generates a noisy predictions where it outputs small positive and negative turn-rates and frequent fluctuations in acceleration. Hence the GAIL-GRU model oscillates within the lane change behaviour Fig. \ref{fig:oscillation_GAIL} (a)-(d); and maneuvers the car off-road Fig. \ref{fig:oscillation_GAIL} (h)-(j); instead of successfully completing the overall task.

In contrast, the proposed MA-GAIL model anticipates the required sub tasks and identifies their sequential relationships via long term planing which led the model to successfully complete the required sub tasks including lane following Fig. \ref{fig:oscillation_MA-GAIL} (a)-(b); lane change to left lane Fig. \ref{fig:oscillation_MA-GAIL} (c)-(e); overtake Fig. \ref{fig:oscillation_MA-GAIL} (f); lane change to right lane Fig. \ref{fig:oscillation_MA-GAIL} (g)-(i) without such oscillations.

We further evaluate the emergent behaviour metric proposed in \cite{kuefler2017imitating} to measure the quality of the demonstrated policy. The considered metrics are 1) lane change rate, 2) off-road duration, 3) hard break rate and 4) traversed distance.

The lane change rate is the average number of times a vehicle makes a lane change within a generated trajectory. Off-road duration is the average number of time steps per trajectory that a vehicle spends more than 1m outside the closest outer road marker. The collision rate is the number of times where the simulated vehicle intersects with another traffic participant. The hard brake rate captures the frequency at which a model chooses to brake harder than -3 $m/s^2$. Finally traversed distance indicates the average total length traversed in the simulated trajectory in kilometres.
 
\begin{table}
  \caption{Emergent behaviour metric evaluations}
  \label{tab:experiment_1}
  \resizebox{.98\linewidth}{!}{
  \begin{tabular}{cccccc}
    \toprule
    Method & Lane change $\downarrow $ & Off-road $\downarrow $  & Hard break $\downarrow $  & Traverse $\uparrow $ \\
    \midrule
    SG 	      &0.66 & 1.20  & 0.31& 0.41 \\
    MR 	      & 0.56 & 0.43 & 0.52& 0.72  \\
    GAIL-GRU & 0.58 & 0.45 & 0.23 & 0.83 \\
    BC 	      & 0.48 & 0.51 & 0.27 & 1.01 \\
    Info-GAIL    & 0.43 & 0.42 & 0.21 & 1.34 \\
    MA-GAIL     & 0.33 & \textbf{0.31} & \textbf{0.19} &  \textbf{1.40}\\
    \hline
    Human        & \textbf{0.31} & 0.43 & 0.20 &  \textbf{1.40}\\
    \bottomrule
  \end{tabular}
  }
	\vspace{-2mm}
\end{table}

The emergent values tabulated in Tab. \ref{tab:experiment_1} shows that the proposed MA-GAIL method is able to outperform all the considered baselines and demonstrates human level control. The SG model performs poorly in all considered methods except hard brake rate because it only drives straight. As a consequence it has a higher collision rate and the smallest traverse distance. We observe an increase of performance from the SG model to MR and the GAIL-GRU model due to the increased capacity of the model to capture dynamic policy of the expert. Still the models perform worse than the BC and Info-GAIL models, largely due to the limited information in the hand-crafted state representation. 

The lack of capacity to model long term temporal dependencies at the sub task level led GAIL-GRU, BC and Info-GAIL models to achieve a higher lane changes and lower traverse distances compared to the MA-GAIL model. In contrast, the proposed MA-GAIL model successfully localises the present context using the local memory and identifies the optimal way to behave using global memory. We would like to further point out that the proposed MA-GAIL model has even outperformed the human expert in the Off-road and Hard break metrics, and matched human level performance in traversed distance metric. 

In Fig. \ref{fig:simulation} we show visual inputs to the proposed MA-GAIL model at different time steps of a particular simulation. The proposed method successfully completes the sub level tasks such as lane change: Fig. \ref{fig:simulation} (a)-(i); over take: Fig. \ref{fig:simulation} (c), (i), (m); turn: Fig. \ref{fig:simulation} (n)-(p); and lane following: Fig. \ref{fig:simulation} (j)-(m); despite the vast diversity of the visual inputs. It should be noted that it demonstrates the lane change from left lane to right Fig. \ref{fig:simulation} (a)-(d) and from right lane to left in Fig. \ref{fig:simulation} (f)-(i). The model possess the capability to understand the current temporal context and has knowledge of different ways it can behave at that particular context. It successfully completes the task at hand and swiftly moves to the next sub task. 

\subsection{Ablation experiments}

To further demonstrate our proposed approach, we conduct a series of ablation experiments identifying the crucial components of the proposed methodology to successfully learn an effective policy. In the same settings as the previous experiment we compare the \textbf{MA-GAIL (proposed)} method to a series of counterparts constructed by removing components of the MA-GAIL model as follows,
\begin{itemize} 
\item \textbf{MA-GAIL / RA}: removes the reward augmentation methodology proposed in Sec. \ref{sec:r_a}. 
\item \textbf{MA-GAIL / } $\mathbf{M^L}$: removes the local memory and retains only the global memory. 
\item \textbf{MA-GAIL / } $\mathbf{M^G}$: removes the global memory and retains only the local memory.
\end{itemize} 

\begin{table}
  \caption{Ablation experiment evaluations}
  \label{tab:experiment_2}
  \resizebox{.98\linewidth}{!}{
  \begin{tabular}{cccccc}
    \toprule
    Method & Lane change $\downarrow $ & Off-road $\downarrow $ & Hard break $\downarrow $  & Traverse $\uparrow $ \\
    \midrule
    MA-GAIL / $M^L$     & 0.59 &0.42 &0.23 &1.03  \\
    MA-GAIL / $M^G$    & 0.41& 0.42 & 0.20 &  1.09\\
    MA-GAIL / RA           & 0.38& 0.38 & 0.21 &  1.28\\
    MA-GAIL (proposed) & \textbf{0.33} & \textbf{0.31} & \textbf{0.19} &  \textbf{1.40} \\
    \bottomrule
  \end{tabular}
  }
	\vspace{-2mm}
\end{table}

The results of our ablation experiment are presented Tab. \ref{tab:experiment_2}. Model MA-GAIL / $M^L$ performs poorly due to it's inability to capture the temporal context of the trajectory and results in frequent lane changes and hard break rates. With a local memory module (i.e MA-GAIL / $M^G$) the oscillations are reduced compared to MA-GAIL/$M^{L}$ as the model can clearly identify the transition between sub tasks. The comparison between models MA-GAIL / RA and MA-GAIL (proposed) clearly emphasises the importance of reward augmentation. The transition between sub level tasks are even smoother in the MA-GAIL (proposed) method (i.e lower Hard break and Lane change rates) as the method clearly identifies the series of sub tasks at hand and achieves them optimally using the experiences stored in $M^L$ and $M^G$, and tries to minimise diverse state transitions as much as possible.

We would like to further compare evaluation results in of Tab. \ref{tab:experiment_2} to those in Tab. \ref{tab:experiment_1} where we observe lower hard break rates and off-road distances compared to all the baseline models considered. This is due to the fact that the MA-GAIL model still has the ability to capture the basic dynamics in driving, even with only a single memory module.  

\section{Conclusions}

In this paper we propose a method to imitate complex human strategies, properly analysing their temporal accordance at a sub task level and identifying strategic differences and similarities among expert demonstrations. We extend the standard GAIL framework with the ability to oversee the history of a long term task, localise the current subtask being completed, and perform long term planning to achieve the overall task. As the process is data driven, it doesn't require any supervision beyond expert demonstrations and could be directly transferred to different tasks without any architectural alterations. Additionally, we introduced a reward augmentation procedure using memory hidden states for smoothing the state transitions, eradicating sudden undesirable manoeuvers in the generated policy. Our quantitative and qualitative evaluations in the TORCS simulation platform clearly emphasise the capacity of the proposed MA-GAIL method to learn complex real world policies and even out performs the human demonstrators.  

\subsection*{Acknowledgement}
\small{
\vspace{-1mm}
This research was supported by an Australian Research Council's Linkage grant (LP140100221). The authors also thank QUT High Performance Computing (HPC) for providing the computational resources for this research.
}

%%%%%%%%%%%%%%%%%%%%%%%%%%%%%%%%%%%%%%%%%%%%%%%%%%%%%%%%%%%%%%%%%%%%%%%%%%%%%%%%%%%%%%%%%%%%%%%%%%%%%%%%%
%% bibliography: see CFP for number of permitted pages

\bibliographystyle{ACM-Reference-Format}  % do not change this line!
\bibliography{sample-bibliography}  % put name of your .bib file here

\end{document}